\documentclass[journal]{IEEEtran}
\usepackage{graphicx}
\usepackage{amsmath}
\usepackage{amssymb}
\usepackage{color}
\usepackage{multirow}
\usepackage{subfig}
\usepackage{bm}
\usepackage{float}

\usepackage[noadjust]{cite}

\begin{document}

\title{A Two-stream Neural Network for Pose-based Hand Gesture Recognition}

\author{Chuankun Li,
        Shuai Li,
        Yanbo Gao,        
        Xiang Zhang,
        Wanqing Li,~\IEEEmembership{Senior Member,~IEEE}
\thanks{
 Chuankun Li is with the North University of China, Taiyuan, China (email: chuankun@nuc.edu.cn).\par Shuai Li and Yanbo Gao are with Shandong University. (e-mail: \{shuaili, ybgao\}@sdu.edu.cn) \par
Xiang Zhang is with University of Electronic Science and Technology of China, Chengdu, China. (e-mail: uestchero@uestc.edu.cn) \par Wanqing Li is with the Advanced Multimedia Research Lab, University of Wollongong, Wollongong, Australia. (wanqing@uow.edu.au).}}
\maketitle


\begin{abstract}
Pose based hand gesture recognition has been widely studied in the recent years. Compared with full body action recognition, hand gesture involves joints that are more spatially closely distributed with stronger collaboration. This nature requires a different approach from action recognition to capturing the complex spatial features. Many gesture categories, such as ``Grab'' and ``Pinch'', have very similar motion or temporal patterns posing a challenge on temporal processing. To address these challenges, this paper proposes a two-stream neural network with one stream being a self-attention based graph convolutional network (SAGCN) extracting the short-term temporal information and hierarchical spatial information, and the other being a residual-connection enhanced bidirectional Independently Recurrent Neural Network (RBi-IndRNN) for extracting long-term temporal information. The self-attention based graph convolutional network has a dynamic self-attention mechanism to adaptively exploit the relationships of all hand joints in addition to the fixed topology and local feature extraction in the GCN. On the other hand, the residual-connection enhanced Bi-IndRNN extends an IndRNN with the capability of bidirectional processing for temporal modelling. The two streams are fused together for recognition. The Dynamic Hand Gesture dataset and First-Person Hand Action dataset are used to validate its effectiveness, and our method achieves state-of-the-art performance.

\end{abstract}


\section{\textsc{Introduction}}

Hand gesture recognition (HGR) is a hot research topic in artificial intelligence and computer vision, and has a large number of applications in the human-machine systems~\cite{Cheng2019ARS, Xue2019MultimodalHH, Yang2020PerformanceCO, 7208833}. Nowadays, depth sensors, like Intel RealSense and Microsoft Kinect, have become easily accessible, so have the body and hand skeletons \cite{Li2019DeepMS,Cao2019SkeletonBasedAR, Song2017AnES}. Accordingly, hand gesture recognition based on pose/skeleton has attracted more and more interests\cite{de2016skeleton,de2017shrec,ohn2013joint,chen2017motion, devineau2018deep,devanne20153}. In the past few years, hand gesture recognition based on handcrafted features has been widely reported~\cite{de2016skeleton,ren2011robust, Wang2012HandPR, marin2014hand}. Generally, shape of connected joints, hand orientation, surface normal orientation are often used as spatial features, and dynamic time warping or hidden Markov model are then used to process temporal information for classification.

Recently, deep learning has also been explored for hand gestures recognition.~\cite{chen2017motion,zhang2017learning, devineau2018deep,shin2020skeleton-based,narayana2018gesture}. One approach is to encode joint sequences into texture images and feed into Convolutional Neural Networks (CNNs) in order to extract discriminative features for gesture recognition. Several methods~\cite{devineau2018deep,narayana2018gesture, molchanov2016online,nguyen2019a} have been proposed along this approach. However, these methods cannot effectively and efficiently express the dependency between joints, since hand joints are not distributed in a regular grid but in a non-Euclidean domain. To address this problem, graph convolutional networks (GCN)~\cite{li2019spatial,nguyen2019a, Hu2020ProgressiveRL} expressing the dependency among joints with a graph have been proposed. For instance, the method in~\cite{li2019spatial} sets four types of edges to capture relationship between non-adjacent joints. However, the topology of the graph is fixed which is ineffective to deal with varying joint relationships for different hand gestures. A typical example is that, the connection between tip of thumb and tip of forefinger in gesture ``Write" is likely to be strong, but it is not the case for gestures ``Prick" and ``Tap". Modelling gesture dependent collaboration among joints is especially important for robust hand gesture recognition. Furthermore, conventional ST-GCN has limited receptive field in temporal domain, hence, long-term temporal information cannot be effectively learned.


Another approach is to apply Recurrent Neural Networks (RNNs) to a hand skeleton sequence for classification~\cite{chen2017motion, shin2020skeleton-based, zhang2017learning}. However, due to the problem of gradient exploding and vanishing in the temporal direction or gradient decay along layers in training a RNN, shallow RNNs are often adopted. This paper adopts the Independently Recurrent Neural Network (IndRNN)~\cite{li2018independently, Li2019DeepIR} as the basic component to develop a deep residual bidirectional IndRNN (RBi-IndRNN) for effective extraction of long-term temporal features.

The contributions are summarized in the following.

\begin{itemize}
\item A two-stream neural network is proposed. One stream is a self-attention based graph convolutional network (SAGCN) and the other is a deep residual-connection enhanced bidirectional Independently Recurrent Neural Network (RBi-IndRNN).

\item The SAGCN is specifically designed to model the strong collaboration among joints in hand gestures. In particular, a global correlation map is first adaptively constructed using a self-attention mechanism to characterize pairwise relationships among all the joints and it works together with a static adjacency map of the hand skeleton to capture varying spatial patterns for different gestures.

\item The deep residual bidirectional IndRNN (RBi-IndRNN) is developed, which extends the residual connection enhanced IndRNN with bidirectional processing to exploit the bidirectional temporal context and long-term temporal information for challenging gestures having similar motion patterns such as ``Grab'' vs ``Pinch'', ``Swipe Left'' vs ``Swipe Right''. The proposed RBi-IndRNN compensate the inability of the SAGCN in learning long-range temporal patterns.

\item Extensive experiments have been performed on two datasets including Dynamic Hand Gesture dataset (DHG)~\cite{de2016skeleton} and First-Person Hand Action datasets (FPHA)~\cite{FirstPersonAction_CVPR2018} and the proposed method performs the best. Detailed analysis on the experimental results is also presented to provide insight on the proposed two-stream networks.
\end{itemize}


Part of this work has been presented in~\textit{ISCAS 2020} \cite{shuai2020iscas} where the method only consists of a plain Bi-IndRNN. This paper extends the method in \cite{shuai2020iscas} to a two-stream network. One stream is a self-attention based graph convolutional network for capturing spatial and short-term temporal information, and the other stream is the deep residual-connection enhanced Bi-IndRNN to exploit the long-term temporal pattern. More experiments are also conducted with thorough analysis.

The rest of this article is structured as follows. Section~\ref{sec:RELATED WORK} provides a review of the relevant literature. The proposed two-stream network is described in Section~\ref{sec:APPROACH}. Extensive experiment and detailed analysis are given in section~\ref{sec:EXPERIMENT}. Section~\ref{sec:CONCLUSION} concludes the paper.



\section{\textsc{Related Work}}
\label{sec:RELATED WORK}
Some related and representative works on hand gesture recognition are introduced in this section. They can be categorized into two approaches: handcrafted features-based~\cite{ren2011robust,Wang2012HandPR,Kuznetsova2013RealTimeSL,marin2014hand,de2016skeleton} and deep learning-based~\cite{chen2017motion,devineau2018deep,narayana2018gesture,zhang2017learning,molchanov2016online,liu2017continuous}.

\subsection{Handcrafted Feature based Hand Gesture Recognition}
Under the category of handcrafted features based methods, Ren et al.~\cite{ren2011robust} modelled hand skeleton as a signature and proposed FEMD (finger-earth mover distance) metric treating each finger as a cluster and penalizing the empty finger-hole to recognize the hand gesture. Wang et al.~\cite{Wang2012HandPR} proposed a mapping score-disparity cost map as the hand representation and used a SVM classifier for recognition. Kuznetsova et al.~\cite{Kuznetsova2013RealTimeSL} used some geometric features to characterize the hand skeleton such as point distances and angle between the two lines created by two points. The multi-layered random forest is used for classification. Marin et al.~\cite{marin2014hand} constructed a feature set containing the position and direction of fingertips, and performed classification using multi-class SVM classifier for recognition of hand gestures. Smedt et al.~\cite{de2016skeleton} designed features based on hand shape and combined featues of wrist rotations and hand directions, and utilized temporal pyramid to model these features.

\subsection{Deep Learning based Hand Gesture Recognition}
In recent years, deep learning based methods have been widely studied for recognizing hand gesture. Depending on the types of deep neural networks, they can be usually divided into two ways: RNNs-based and CNNs-based ones.

\begin{figure*}[tb]
\centering
\includegraphics[width=0.9\textwidth,height=0.3\textwidth]{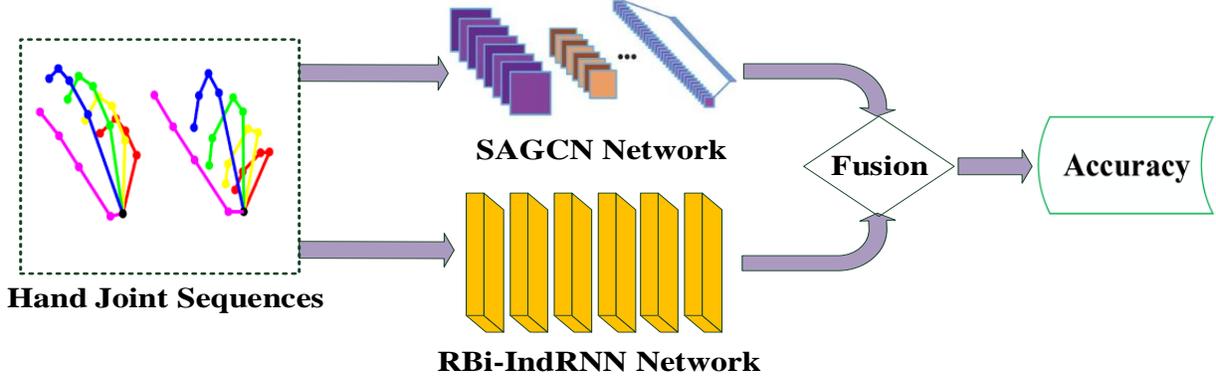}
\caption{Framework of the proposed two-stream network. SAGCN module focuses on hierarchical spatial information and short-term temporal information, and RBi-IndRNN focuses on long-term temporal information.}
\label{fig_streamframework}
\end{figure*}

For the CNNs-based approach, it takes the advantage of CNNs to process spatial hand joints or each hand joint over time as grid data and extract local features. Devineau et al.~\cite{devineau2018deep} employed a parallel CNN to conduct temporal convolution processing for the world coordinate sequence of the skeleton joint separately, and fused the time-varying characteristics of each joint into a full connection layer (FC) and Softmax function for hand gesture classification.
 Liu et al.~\cite{liu2017continuous} firstly used RGB and depth map based faster region proposal networks to detect hand region. Then the video was segmented by using hand positions. From segmented videos, hand-oriented spatio-temporal features were extracted using the 3D convolutional networks for recognizing the hand gestures.
 Narayana et al.~\cite{narayana2018gesture} utilized a spatial focus of attention to construct 12-stream networks called a FOA-Nets. Each network extracted local features from different modality. Then a sparse network architecture was designed to fuse the 12 channels.
 Pavlo et al.~\cite{molchanov2016online} used a recurrent 3DCNN for hand gesture recognition. In the recurrent 3DCNN, connectionist temporal classification (CTC) was utilized to synchronously segment video  and classify dynamic hand gestures from multi-modal data.
  However, these methods based on CNNs cannot utilize the graphical structure of hand joints explicitly. To exploit this, Li et al.~\cite{li2019spatial} employed a hand gesture based graph convolutional network (HGCN) to capture the linkage and motion information of different joints. However, the topology of the graph is fixed which limits capability of the network. In addition, the GCN is not designed for learning long-term temporal relationship. Nguyen et al. \cite{nguyen2019a} employed a neural network to exploit gesture representation from hand joints by using SPD matrix. This method focuses on relationship of local joints, but ignores two joints that are far away from each other in the matrix. Motivated by the concept of attention based graph neural network (GCN) \cite{Guo2019AttentionBS, Shi2019TwoStreamAG}, we propose a self-attention based GCN (SAGCN) to effectively explore the often strong collaboration among hand joints in gesture.

  %


  In the RNNs-based approach, handcrafted features or features extracted from CNN of joint sequences  are used as input to RNNs to explore temporal information.
Chen et al.~\cite{chen2017motion} utilized angle features of joints to characterize the finger movement and global features of rotation and translation. Three LSTM networks were used to process these features, respectively. Shin et al.~\cite{shin2020skeleton-based} used two local features (finger and palm features) and a global feature (Pose feature) to capture the spatial information, and used a GRU-RNN network to process each feature part. Zhang et al.~\cite{zhang2017learning} employed a deep framework including 3DCNN layer, ConvLSTM layer, 2DCNN layer and temporal pooling layer to capture short-term spatio-temporal information.
 However, due to the difficulty in constructing a deep RNN, these methods usually employ a relatively shallow RNN network. Consequently, they often fail to recognize hand gesture having complex and long-term temporal patterns. In this paper, a residual connection enhanced bidirectional Independently Recurrent Neural Network (RBi-IndRNN) is employed to address this issue.

\section{\textsc{Proposed Method}}
\label{sec:APPROACH}

Fig. \ref{fig_streamframework} illustrates the proposed two-stream network, one stream is a self-attention based Graph Convolutional Network learning the spatial and short-term temporal features, and the other stream is a residual Bi-IndRNN learning long-term temporal features. The two streams are then fused at score level for final classification.

\begin{figure}[tb]
  \centering
  \includegraphics[width=3.4in]{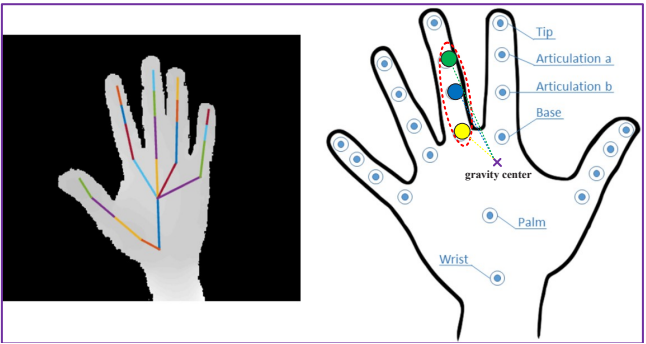}
  \caption{Sample of hand skeleton in the DHG 14/28 dataset~\cite{de2016skeleton}. Referencing to the gravity center of the hand joints (purple cross), the neighbouring connected joints of each joint, called root (blue circle), are grouped into two: 1) centripetal group (yellow circle): adjacent nodes closer to the gravity center of the hand joint; 2)centrifugal group: the rest connected neighbors (green circle). }
  \label{fig_skeleton}
\end{figure}

\subsection{Self-attention based Graph Convolutional Network}
Graph Convolutional Networks (GCNs) have been proposed for structured but non-grid data and extended for recognizing human action based on skeleton. A typical example is the spatial temporal GCN (ST-GCN)~\cite{yan2018spatial}. However, ST-GCN mainly explores the interaction or collaboration between local (e.g. first and second order neighbouring) joints and the topology of the graph representing the human body is fixed or static. Such representation is ineffective in exploring the collaboration among non-neighbouring joints and, hence, in differentiating actions with collaboration of different body parts. This problem becomes severe in hand gesture recognition where hand joints, regardless of being connected or not, are more closely and strongly collaborated than body joints in action recognition. To tackle this problem, a self-attention based ST-GCN method is developed.

As illustrated in the left image of Fig.~\ref{fig_skeleton}, a hand skeleton graph is usually defined with each hand joint being a node and anatomical connectivity between joints being edges. The graph is expressed as an adjacency matrix whose elements representing connected neighbouring joints are one and, otherwise, zero. Convolutional operations are performed at each node, the node itself, referred to as a root node to differentiate it from its connected nodes in the following, and its neighbouring connected nodes defined by the adjacency matrix. In order to characterize the local structure, different weights are learned for the root node and its connected nodes. Since the number of its connected nodes varies from node to node, a spatial configuration partitioning strategy is developed for hand joints similarly as in \cite{yan2018spatial} to generalize the operations by defining two groups of neighbouring connected nodes and assuming the same weight for nodes in the same group. Specifically, a gravity center is first generated by calculating the average coordinate of all hand joints in a frame. Using this gravity center as a reference point, the neighbouring adjacent nodes/joints of each root node are grouped into two: 1) centripetal group: adjacent nodes closer to the gravity center of the hand joint; 2) the centrifugal group: the other neighbouring nodes locating farther away from the gravity center than the root node/joint, as illustrated the right image in Fig.~\ref{fig_skeleton}. With the above adjacency matrix and partitioning strategy, a convolution operation on each node can be expressed as follows.

\begin{equation}
\mathbf{f_{out}} = \sigma(\sum_{k}^{K_{v}}\mathbf{W_{k}f_{in}}\mathbf{A_{k}})
\end{equation}
\noindent where $K_{v}$ is the kernel size of the spatial dimension, i.e., the number of groups of connected neighbors, and is set to 3 (the two groups plus the root). For brevity, $\mathbf{A_k}$ denotes the normalized adjacency matrix $\mathbf{\Lambda_k ^{-\frac{1}{2}}(A_k)\Lambda_k ^{-\frac{1}{2}}}$ similarly as in\cite{Shi2019TwoStreamAG}, where $\Lambda_k^{ii}=\sum_j(A_k^{(ij)})$. $\mathbf{W_{k}}$ is the weight vector of the $1\times1$ convolution operation.  The nonlinear activation function ReLU is used for $\sigma$.


The above adjacency matrix is predefined anatomically and static. It represents each joint and its connected neighbouring joints, hence, the convolution operation extracts local features. However, such a local process is not able to capture directly the collaboration among non-connected joints that is often dynamic within a gesture and varies from gesture to gesture. Therefore, we propose in this paper a dynamic attention matrix or map to adaptively characterize the collaboration among all nodes/joints. The attention matrix is obtained via a self-attention mechanism and is calculated in a feature space by projecting the node features with a weight $\mathbf{W_a}$. The attention matrix is especially useful for the spatially closely-located, connected or non-connected hand joints that have strong collaboration. The detailed process is expressed as follows:

\begin{align}
\mathbf{f_{a}} &= \mathbf{W_a  {f_{in}}} \\
\mathbf{A_g} &= \frac{exp(\mathbf{f_{a}}\otimes \mathbf{f_{a}^{T}})}{\sum exp(\mathbf{f_{a}}\otimes \mathbf{f_{a}^{T}})}
\label{eq_att}
\end{align}

\noindent where $\mathbf{f_{in}}$ is input and $\mathbf{W}$ maps the input to a feature space (done via a convolutional operation in the experiments). $\mathbf{f_{a}^{T}}$ is the transpose of $\mathbf{f_{a}}$, and $\otimes$ is matrix multiplication. $\mathbf{A_g}$ is an attention matrix to indicate the relationship of the pair-wise nodes/joints and is normalized to $(0, 1)$ via softmax.

In order to capture local structure and global collaboration among the joins together, $\mathbf{A_g}$ is combined with the adjacency matrix as follows.

\begin{equation}
\mathbf{f_{out}} = \sigma(\sum_{k}^{K_{v}}\mathbf{W_{k}f_{in}}\mathbf{A_{k}}+\mathbf{W_{g}f_{in}}\mathbf{A_g})
\end{equation}
\noindent where $\sum_{k}^{K_{v}}\mathbf{W_{k}f_{in}}\mathbf{A_{k}}$ captures local structure of joints and their connected neighbors and $\mathbf{W_{g}f_{in}}\mathbf{A_g}$ captures the global collaboration among all the joints. In this way, the proposed self-attention based GCN, termed as SAGCN, can process both local and global features together. Finally, the spatial features at each time step are processed over time with convolution in the same way as TCN in \cite{yan2018spatial}. 

\begin{figure}[tb]
\centering
\includegraphics[width=0.5\textwidth,height=0.17\textwidth]{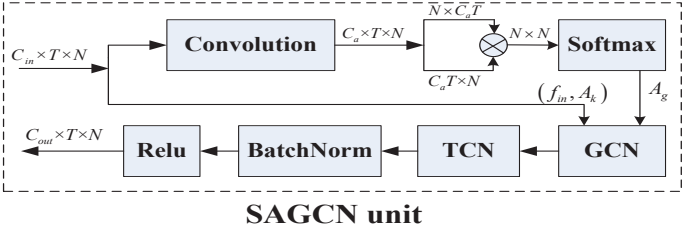}
\caption{Illustration of an SAGCN unit. }
\label{fig_SAGCNf}
\end{figure}
\begin{figure}[tb]
\centering
\includegraphics[width=0.5\textwidth,height=0.1\textwidth]{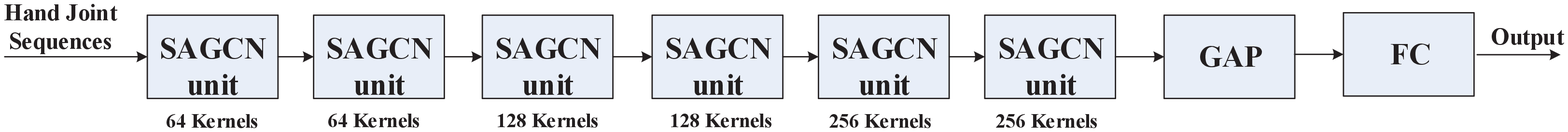}
\caption{The recognition network built upon the proposed SAGCN. }
\label{fig_SAGCNf_framework}
\end{figure}

Fig.~\ref{fig_SAGCNf} illustrates the proposed SAGCN. An input feature is first processed by a convolution operation to a feature space, and then multiplied and normalized with a Softmax function to obtain the attention map $A_g$. Together with the input adjacency matrix, the GCN and TCN process the input feature spatially and temporally with batch normalization and ReLU activation functions. $N$, $C$ and $T$ in the figure represent the total number of the vertexes, number of convolutional channels and the length of hand sequences, respectively. The structure of the proposed SAGCN used in the experiments is illustrated in Fig. \ref{fig_SAGCNf_framework}. The network contains six layers of SAGCN units and in each SAGCN layer the numbers of kernels of the attention component, the GCN component and the TCN component are set the same. The numbers of kernels for the six SAGCN layers are 64, 64, 128, 128, 256, and 256, respectively. The stride for the convolution in the fourth TCN is set to $2$ as a pooling operation over time. After six layers of SAGCN units, global average pooling (GPA) is used to pool the spatial-temporal features, and feed into FC and Softmax function for gesture classification.

\begin{figure}[tb]
\centering
\includegraphics[width=3.2in]{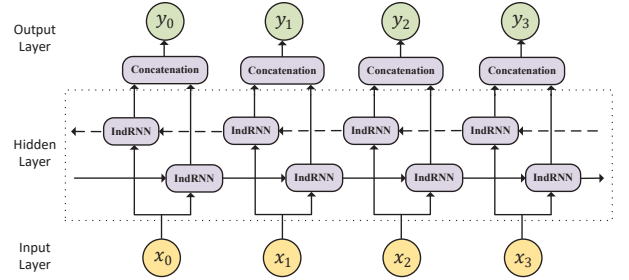}
\caption{Structure of the proposed bidirectional IndRNN.}
\label{fig:biIndrnn}
\end{figure}

\subsection{Residual Bidirectional Independently Recurrent Neural Network (RBi-IndRNN)}
As seen from the architecture of an SAGCN unit, it focuses on hierarchical spatial features and relatively short-term temporal features since it has a relatively small receptive field of the temporal convolution operations. It would be difficult for an SAGCN to capture long-term temporal features, hence, to distinguish gestures that are differentiated in long-term motion patterns, such as ``Swipe up'' and ``Swipe down''. To address this shortcoming of the SAGCN, a second stream using the recent IndRNN \cite{li2018independently, Li2019DeepIR} is proposed in this paper.

\begin{equation}
\mathbf{h}_t=\sigma(\mathbf{Wx}_t+\mathbf{u}\odot\mathbf{h}_{t-1}+\mathbf{b})
\end{equation}
where $\mathbf{x}_t\in \mathbb{R}^{M}$ is input of IndRNN network and $\mathbf{h}_t\in \mathbb{R}^{N}$  represents hidden states at time-step $t$. $\mathbf{W}\in \mathbb{R}^{N\times M}$, $\mathbf{u}\in \mathbb{R}^{N}$ and $\mathbf{b}\in \mathbb{R}^{N}$ are the weights need to be learned. $\odot$ is dot product, and $N$ is the number of neurons. One of the key advantages of IndRNN, comparing with the conventional RNN, is that IndRNN can capture longer temporal information by regulating the recurrent weights to avoid gradient vanishing and exploding in training. Moreover, multiple IndRNN layers are able to be efficiently stacked to build a deeper network with low complexity.

\begin{figure}[tb]
\centering
\subfloat[ ]{
\label{fig:improved_resindrnn_1}
\begin{minipage}[t]{0.4\linewidth}
\centering
\includegraphics[height = 48mm, width = 40mm]{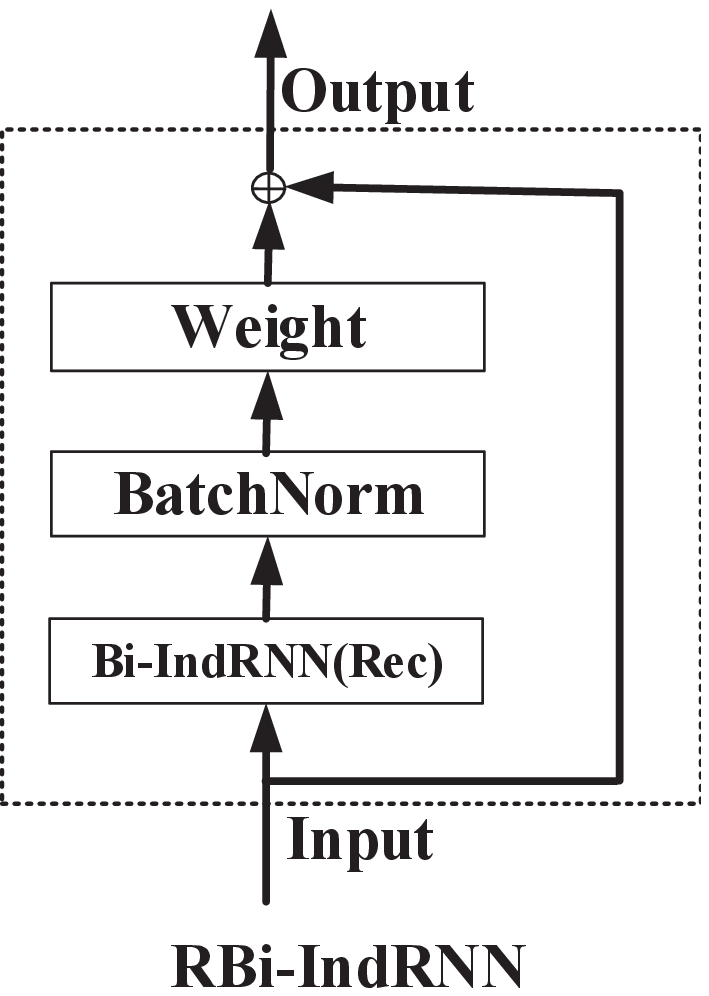}
\end{minipage}
}
\subfloat[ ]{
\label{fig:6layerbiIndrnn}
\begin{minipage}[t]{0.4\linewidth}
\centering
\includegraphics[height = 60mm, width = 25mm]{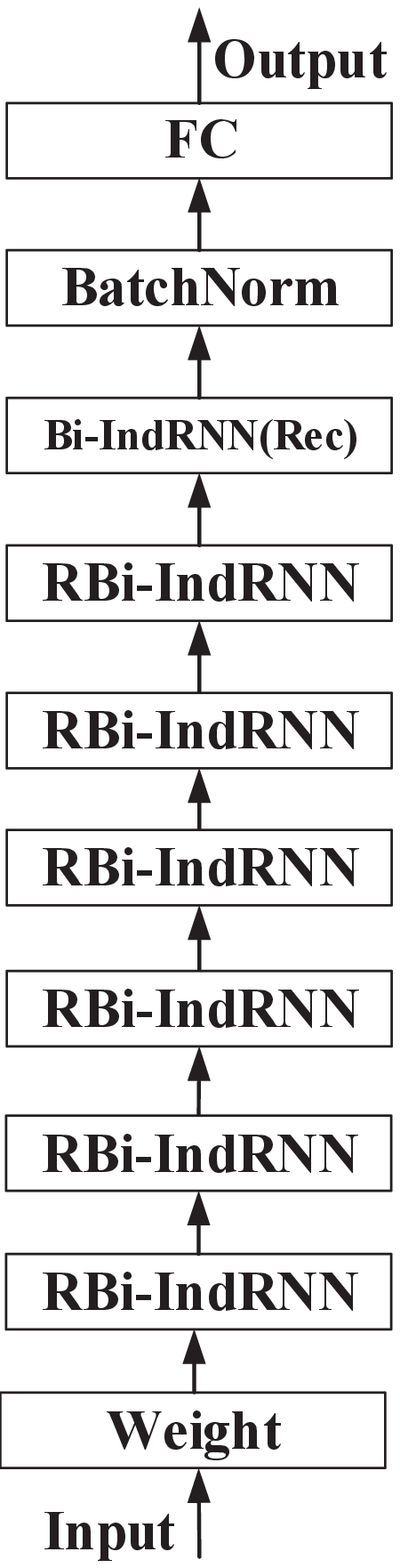}
\end{minipage}
}

\caption{Illustration of the (a): RBi-IndRNN module and (b): proposed 6-layer RBi-IndRNN for hand gesture recognition. }
\label{fst}
\end{figure}

Considering the success of the Bidirectional Recurrent Neural Networks (Bi-RNN) in action recognition~\cite{liu2018multi} and language modelling~\cite{arisoy2015bidirectional}, a Bidirectional IndRNN (Bi-IndRNN) is constructed whose architecture is shown in Fig.~\ref {fig:biIndrnn}. The features are extracted by two directions of temporal processing using IndRNNs, and features of two directions are concatenated and fed into next layer. The Bi-IndRNN is able to capture the temporal relationship in two directions. Considering that IndRNN can effectively work with ReLU, we further develop a residual connection enhanced Bi-IndRNN (RBi-IndRNN) as shown in Fig.~\ref{fig:improved_resindrnn_1}, by adding an identity shortcut (skip-connection) to bypass the non-linear transformation of the input feature in order to facilitate the gradient backpropagation. This skip-connection does not affect the temporal processing, but makes the deeper features a summation of the shallower features. This paper adopts a pre-activation type of residual function with batch normalization, Bi-IndRNN and then weight processing as shown in Fig.~\ref{fig:improved_resindrnn_1}.


In the experiments, a deep RBi-IndRNN of 6-layers and the framework is shown in Fig.~\ref {fig:6layerbiIndrnn}. After six layers of RBi-IndRNN, the features of last time-step are fed into a FC layer for gesture classification.  Moreover, the temporal displacement of each joint describing the movement between adjacent frames similarly as the optical flow is extracted and concatenated with original skeletal joints as input to the RBi-IndRNN.


\begin{figure*}[tb]
\centering
\includegraphics[width=0.9\textwidth,height=0.3\textwidth]{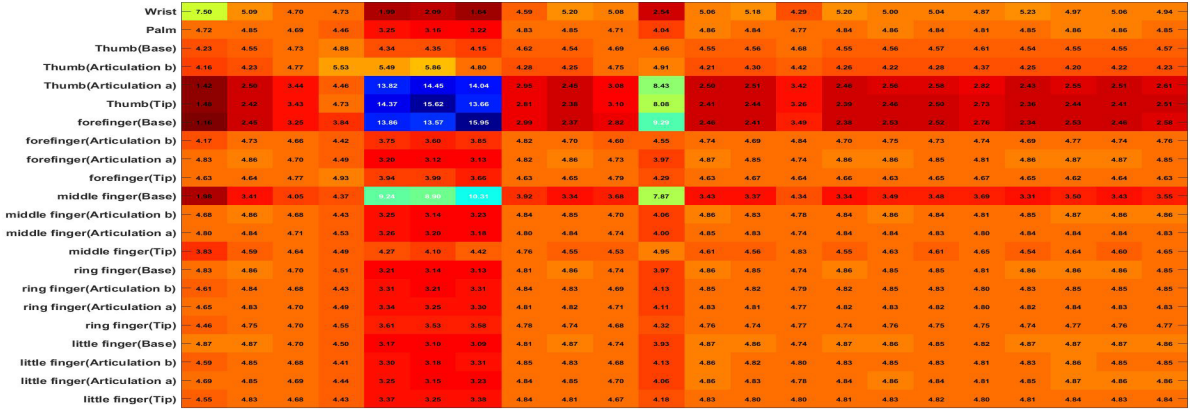}
\caption{A sample attention matrix of gesture ``Grab'' obtained by the proposed SAGCN. }
\label{fig_a}
\end{figure*}

\begin{figure*}[tb]
\centering
\subfloat[ ]{
\label{tab:confusion14}
\begin{minipage}[t]{0.5\linewidth}
\centering
\includegraphics[width = 1\textwidth]{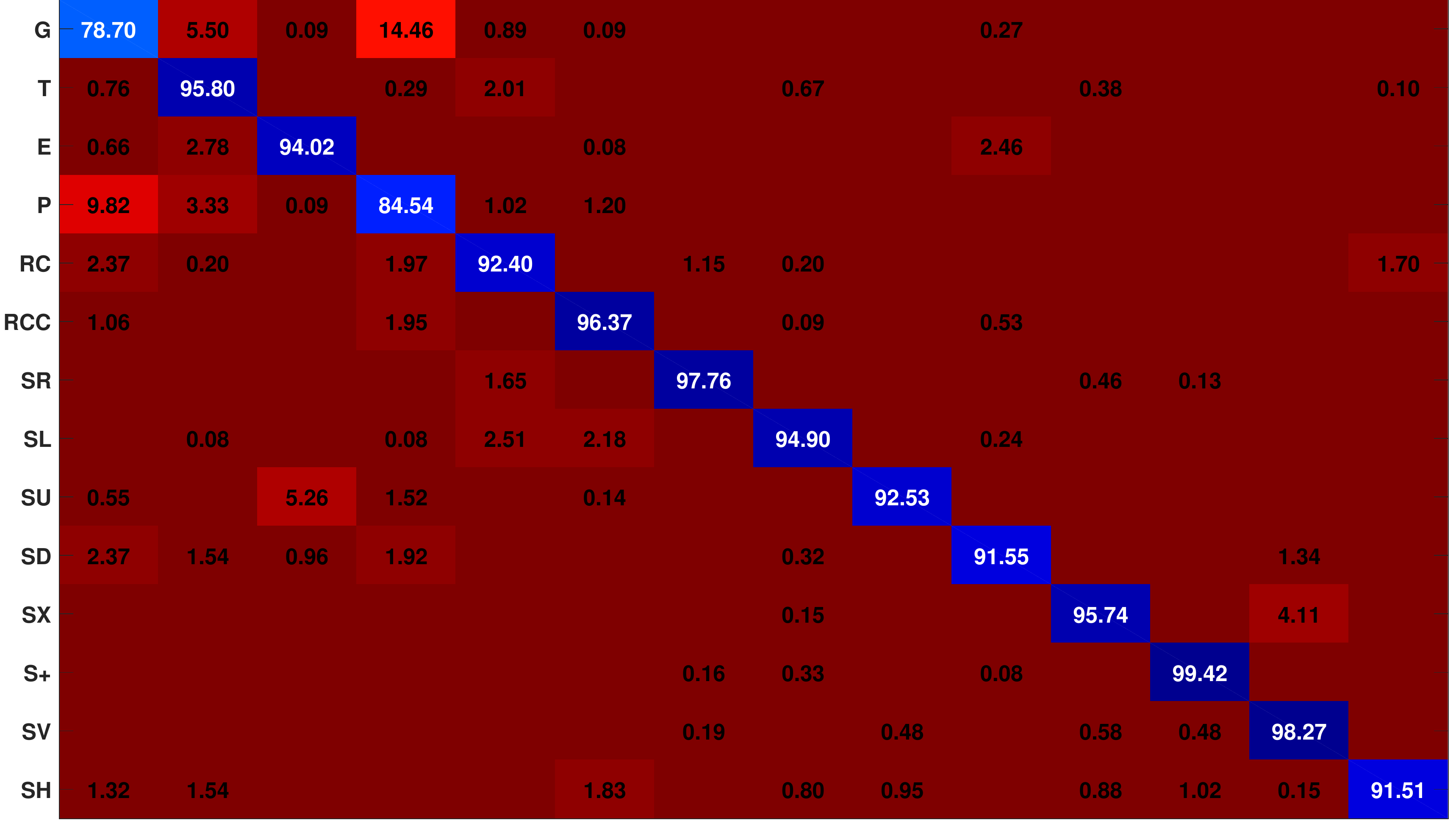}
\end{minipage}
}
\subfloat[ ]{
\label{cnn14}
\begin{minipage}[t]{0.5\linewidth}
\centering
\includegraphics[width = 1\textwidth]{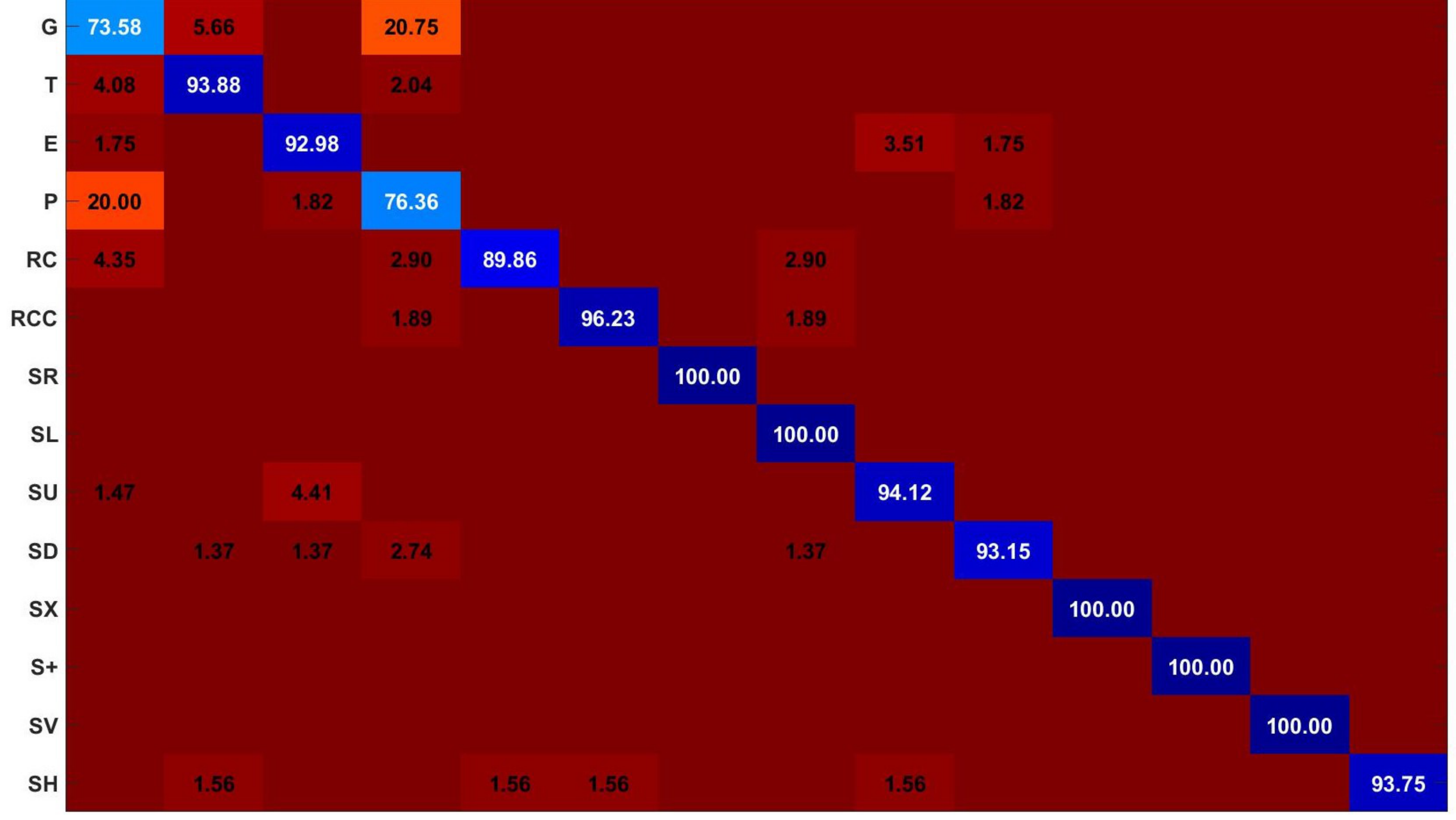}
\end{minipage}
}
\caption{Confusion matrices on the DHG-14 dataset using (a): RBi-IndRNN and (b): SAGCN. }
\label{conf_comp}
\end{figure*}

In order to reduce the complexity of the two-stream network training, the SAGCN and the RBi-IndRNN are trained separately. The resulted models are then used to test and the probability vectors produced by the SAGCN and RBi-IndRNN are fused by multiplication. The class with the max probability is recognized as the gesture class and can be expressed as follows.
 \begin{equation}
 label = argmax({v_{SAGCN}} \odot {v_{RBi-IndRNN}})
 \end{equation}
where $v$ represents a probability vector, $\odot$ is dot product, and $argmax(\cdot)$ is to find the index position with the maximum probability.

\section{\textsc{Experiments}}
\label{sec:EXPERIMENT}
\subsection{Datasets}
Two widely used datasets for HGR, namely the Dynamic Hand Gesture (DHG) 14/28 dataset~\cite{de2016skeleton} and the First-Person Hand Action (FPHA) dataset~\cite{FirstPersonAction_CVPR2018}, are used for experiments. The DHG 14/28 dataset~\cite{de2016skeleton} is captured by Intel RealSense depth cameras containing hand joint data (3-dimension world coordinate ($x$, $y$, $z$)) and depth map sequences as shown in Fig.~\ref{fig_skeleton}. This DHG 14/28 dataset is constructed with 2800 gesture sequences containing 14 gestures performed by 20 subjects. Each sequence ranging from 20 to 50 frames is assigned with a class label. The categories of this dataset include ``Swipe x (SX)'', ``Swipe down (SD)'', ``Rotation counter-clockwise (RCC)'', ``Tap (T)'', ``Rotation clockwise (RC)'', ``Swipe right (SR)'', ``Pinch (P)'', ``Swipe up (SU)'', ``Shake (SH)'', ``Grab (G)'', ``Swipe left (SL)'', ``Swipe v (SV)'',``Expand (E)'', ``Swipe + (S+)''. Depending on the number of fingers used, gestures are classified with either 14 labels or 28 labels. The evaluation protocols in~\cite{devineau2018deep} are adopted, namely 1960 video as training samples and other sequences for testing. And 5\% of the training data is randomly selected for validation. 20 frames are sampled from each sequence and fed into the proposed networks described in Section~\ref{sec:APPROACH} for training and classification.

The FPHA dataset~\cite{FirstPersonAction_CVPR2018} contains 1175 sequences from 45 different gesture classes with high viewpoint, speed, intra-subject variability and inter-subject variability of style, viewpoint and scale. This dataset is captured in 3 different scenarios (kitchen, office and social) and performed by 6 subjects. Compare with DHG 14/28 dataset, FPHA dataset has 21 hand joints and the palm joint is missed. This is a challenging dataset due to the similar motion patterns and involvement of many different objects. The same evaluation strategy in ~\cite{FirstPersonAction_CVPR2018} are used.

\subsection{Training Setup}
The experiments are conducted on the Pytorch platform using a 1070Ti GPU card. Adaptive Moment Estimation (Adam)~\cite{kingma2014adam} is used as optimization function of training network.
The batch size is set to 64 and 32 for DHG and FPHA datasets, respectively. Dropout~\cite{srivastava2014dropout} is set to 0.2 and 0.5 and the initial learning rate is set to $2*10^{-4}$ and  $2*10^{-3}$ for RBi-IndRNN and SAGCN respectively. When the validation accuracy is improved, learning rate is decayed by 10. The number of neurons in each RBi-IndRNN layer is $512$. And 1024 neurons are used in FC of the RBi-IndRNN due to the bidirectional processing.

\begin{table}[tb]
\centering
\caption{Comparison of w/o self-attention for the GCN on two datasets} \label{attention}
\begin{tabular}{l c c c}
  \hline
  \textbf{Method} & \textbf{DHG-14} & \textbf{DHG-28}& \textbf{FPHA}
  \\
  \hline
 GCN & 90.83\% & 87.74\%& 83.48\% \\
 SAGCN & 93.33\% & 91.54\%& 87.83\% \\
  \hline
\end{tabular}
\end{table}

\begin{table*}[htb]
\centering
\caption{Results on the two datasets under different settings of RBi-IndRNN} \label{eva_indrnn}
\begin{tabular}{l c c c}
  \hline
  \textbf{Method} & \textbf{DHG-14} & \textbf{DHG-28}& \textbf{FPHA}
  \\
  \hline
  IndRNN(joint coordinate) & 92.07\% & 85.82\%& 84.52\% \\
  IndRNN(joint coordinate + displacement) & 92.19\% & 88.87\%& 86.78\% \\
 Bi-IndRNN(joint coordinate + displacement) & 93.15\% & 91.13\%& 88.35\% \\
 RBi-IndRNN(joint coordinate + displacement) & 94.05\% & 91.90\%& 88.87\% \\
  \hline
\end{tabular}
\end{table*}

\begin{table*}[!htb]
\centering
\caption{Performance of the proposed method on the DHG dataset with comparisons to the existing methods in terms of accuracy} \label{tab:resultsheet}
\begin{tabular}{ p{8cm} c c c c c }
  \hline
  \textbf{Method}& \textbf{modality} & \textbf{14 gestures} & \textbf{28 gestures}& \textbf{average}
  \\
  \hline
  HO4D Normals~\cite{oreifej2013hon4d}&Depth & 78.53\% & 74.03\%&76.28\% \\
  Motion Trajectories  ~\cite{devanne20153}&Pose & 79.61\% & 62.00\%& 70.80\%\\
  CNN for key frames~\cite{de2017shrec}&Pose & 82.90\% & 71.90\%&77.40\% \\
  JAS and HOG2~\cite{ohn2013joint}&Pose & 83.85\% & 76.53\%&80.19\% \\
  RNN+Motion feature  ~\cite{chen2017motion}&Pose & 84.68\% & 80.32\%& 82.50\%\\
  HoHD+HoWR+SoCJ ~\cite{de2016skeleton}&Pose & 88.24\% & 81.90\%&85.07\% \\
  Parallel CNN ~\cite{devineau2018deep}&Pose & 91.28\% & 84.35\%&87.82\% \\
  HG-GCN~\cite{li2019spatial}&Pose & 92.80\% & 88.30\%&90.55\% \\
  leap motion controller~\cite{avola2019exploiting}&Pose &  \textbf{97.62\%} & 91.43\%& 94.53\%\\
  PB-GRU-RNN~\cite{shin2020skeleton-based}&Pose     & 95.21\% & 93.23\% &94.22\%\\
  \textbf{proposed method} &Pose &96.31\% &  \textbf{94.05\%} &\textbf{95.18\%} \\
  \hline
\end{tabular}
\end{table*}
\begin{figure}[!htb]
\centering
\includegraphics[width=0.48\textwidth,height=0.25\textwidth]{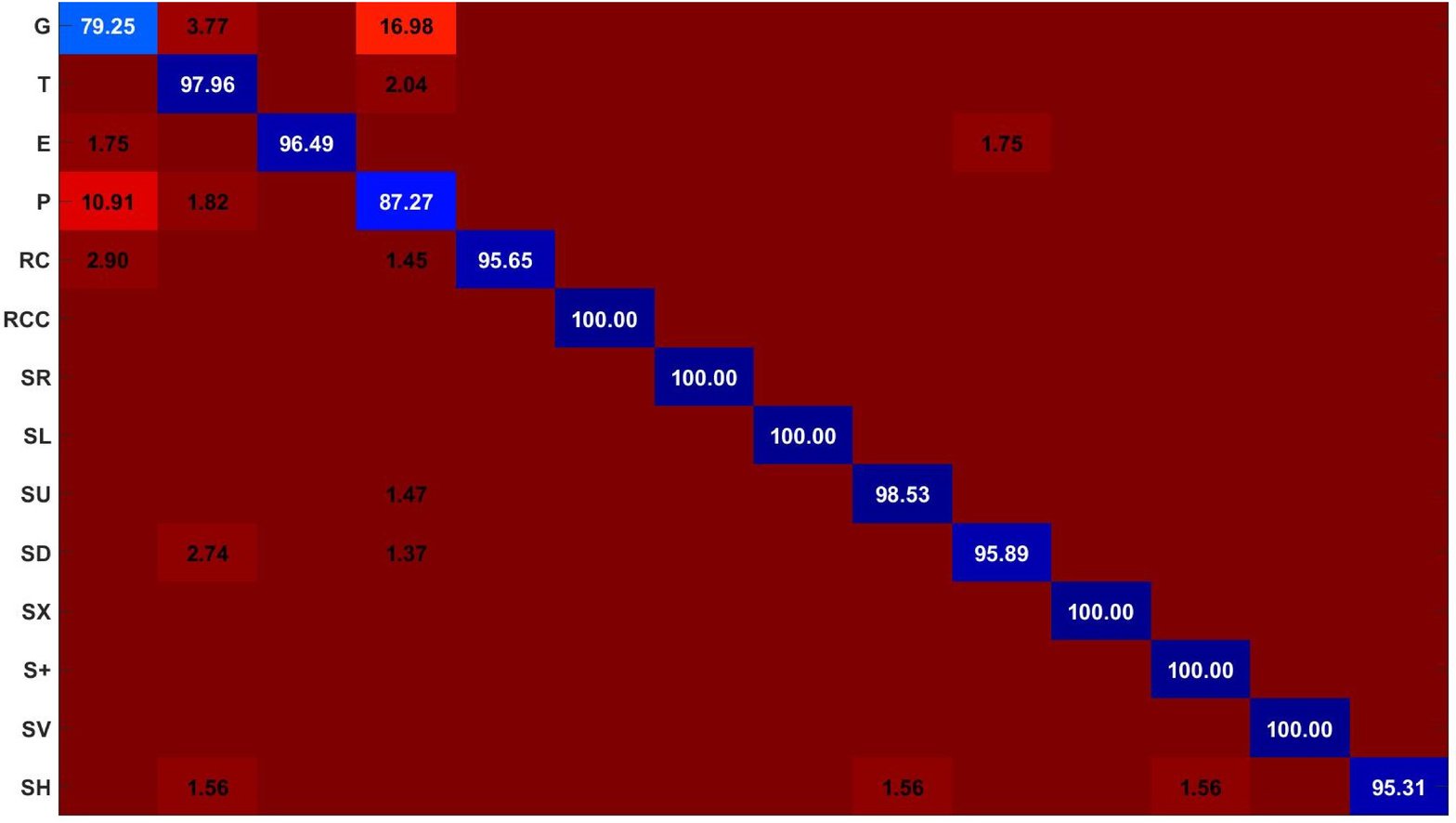}
\caption{Confusion matrix of the proposed network on DHG-14.}
\label{rnn-cnn14}
\end{figure}
\begin{figure*}[!htb]
\centering
\includegraphics[width=0.9\textwidth,height=0.43\textwidth]{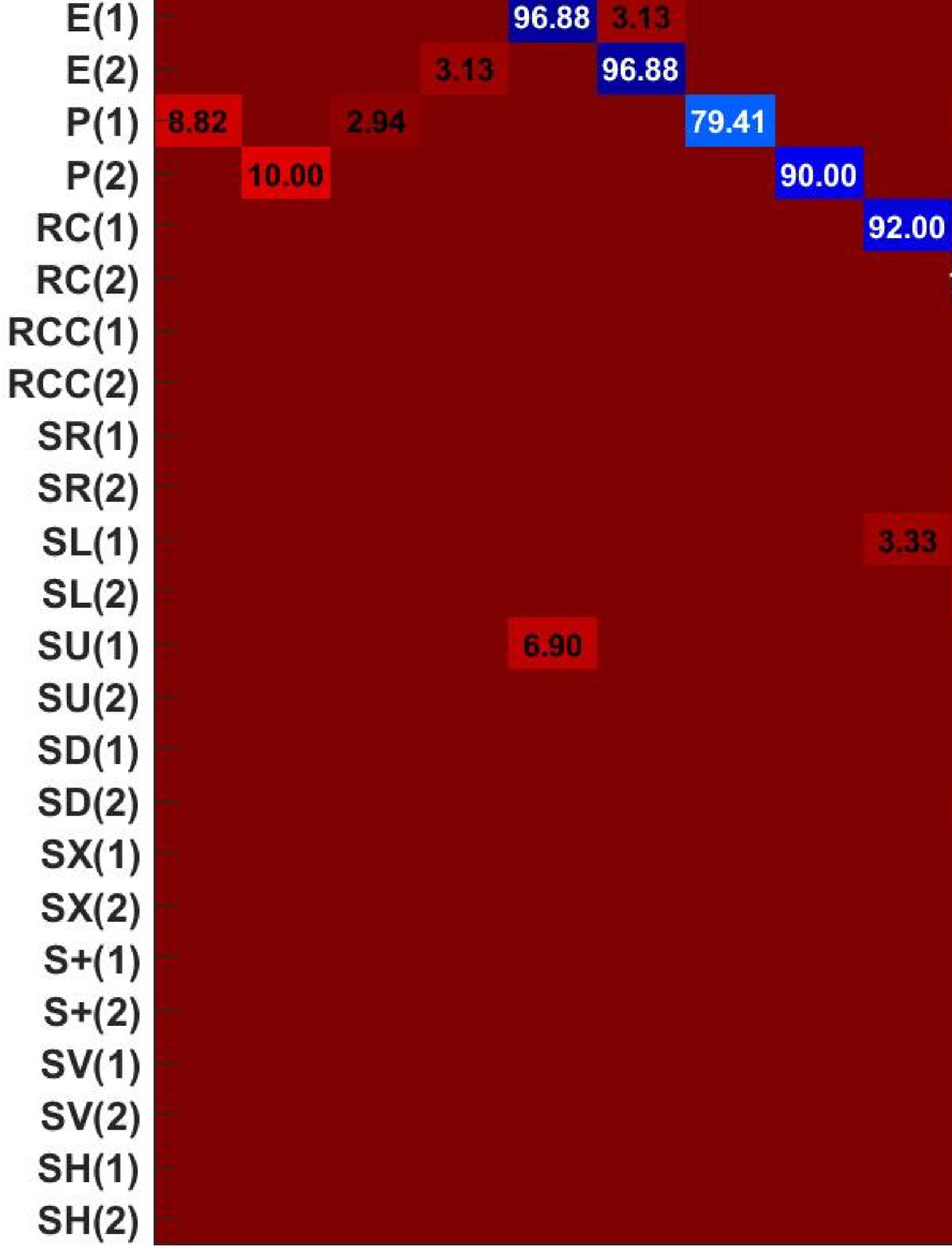}
\caption{Confusion matrix of the proposed network on DHG-28.}
\label{rnn-cnn28}
\end{figure*}

\begin{table}[t]
\centering
\caption{Results on the FPHA dataset and comparisons to the existing methods in terms of the accuracy} \label{FPHAresult}
\begin{tabular}{p{5cm} @{}c@{} @{}c@{} }
  \hline
  \textbf{Method}& \textbf{modality} & \textbf{14 gestures}
  \\
  \hline
  Two stream~\cite{feichtenhofer2016convolutional}&RGB & 75.30\%  \\
  Joint angles similarities and HOG2~\cite{ohn2013joint}& Depth+Pose & 66.78\%  \\
  Histogram of Oriented 4D Normals~\cite{oreifej2013hon4d}&Depth & 70.61\%  \\
  JOULE~\cite{hu2015jointly}& RGB+Depth+Pose     & 78.78\% \\
  Novel View~\cite{rahmani20163d}&Depth & 69.21\% \\
  2-layer LSTM~\cite{zhu2016co-occurrence}&Pose & 80.14\%\\
  Moving Pose~\cite{zanfir2013the}&Pose & 56.34\%  \\
  Lie Group~\cite{vemulapalli2014human}&Pose & 82.69\%  \\
  HBRNN~\cite{du2015hierarchical}&Pose  & 77.40\% \\
  Gram Matrix~\cite{zhang2016efficient}&Pose &  85.39\%\\
  TF~\cite{garciahernando2017transition}&Pose     & 80.69\% \\

 SPD Matrix Learning~\cite{huang2016a}&Pose     & 84.35\% \\
 Grassmann Manifolds~\cite{huang2018building}&Pose     & 77.57\% \\
  \textbf{proposed method} &Pose &90.26\%   \\
  \hline
\end{tabular}
\end{table}

\subsection{Ablation Study on Some Key Factors}

\subsubsection{Contribution of the self-attention in the SAGCN}

The self-attention model used in the proposed SAGCN is evaluated against the GCN. The results on the two datasets by using GCN network and SAGCN are tabulated in Table~\ref{attention}. It shows that the self-attention model used in the proposed SAGCN significantly improves the performance, compared with original GCN network. Taking the FPHA dataset as an example, the performance is improved  from $83.48\%$ to $87.83\%$ in terms of accuracy. It indicates that the collaboration among all hand joints is important and is well captured by the proposed attention model via the attention matrix. A visual illustration, an attention matrix of gesture ``Grab'', is shown in Fig.~\ref{fig_a}. This matrix demonstrates and also verifies the intuition that collaboration between thumb and forefinger joints is very important for recognizing gesture ``Grab''.

\subsubsection{Contribution of the bidirectional processing and residual connection in the RBi-IndRNN}

The bidirectional processing and residual connection in the RBi-IndRNN is evaluated against the plain IndRNN. Table~\ref{eva_indrnn} shows the recognition results under different settings. The classification accuracies of the plain IndRNN only using joint world coordinates are 92.07\%,  85.82\% and 84.52\% on DHG14, DHG-28 and FPHA datasets, respectively, which outperforms some methods (as illustrated in the following Table \ref{FPHAresult} and Table \ref{tab:resultsheet}). When using the temporal displacement and joint world coordinates, the performance is further improved to 92.19\%, 88.87\% and 86.78\%, respectively. It improves by $3.05$ and $2.26$ percentage points for DHG-28 and FPHA datasets respectively, compared with only using joint coordinates. Meanwhile, the RBi-IndRNN with the temporal displacement performs better than Bi-IndRNN and gets the best performance as shown in Table~\ref{eva_indrnn}.

\subsubsection{Contribution of the long-term temporal feature processing using RBi-IndRNN against SAGCN}

As discussed in Section \ref{sec:APPROACH}, SAGCN is not capable of learning long-term temporal features, but the RBi-IndRNN is. By comparing the performance of SAGCN and RBi-IndRNN as illustrated in Table~\ref{attention} and Table~\ref{eva_indrnn}, respectively, RBi-IndRNN outperforms the SAGCN. To further understand what types of hand gestures that SAGCN and RBi-IndRNN are better at in the proposed framework, the confusion matrices of the SAGCN and RBi-IndRNN recognition results on the DHG-14 dataset are presented in Fig.~\ref{tab:confusion14} and \ref{cnn14}, respectively. By comparing the two confusion matrices, it can be seen that SAGCN performs better for hand gestures having much spatial variation and less temporal movement such as `` Swipe left ", while RBi-IndRNN  performs better for hand gestures with complex temporal motion such as ``Grab ". This indicates that SAGCN and RBi-IndRNN are complementary to each other.

\begin{figure*}[!htb]
\centering
\includegraphics[width=0.95\textwidth,height=0.39\textwidth]{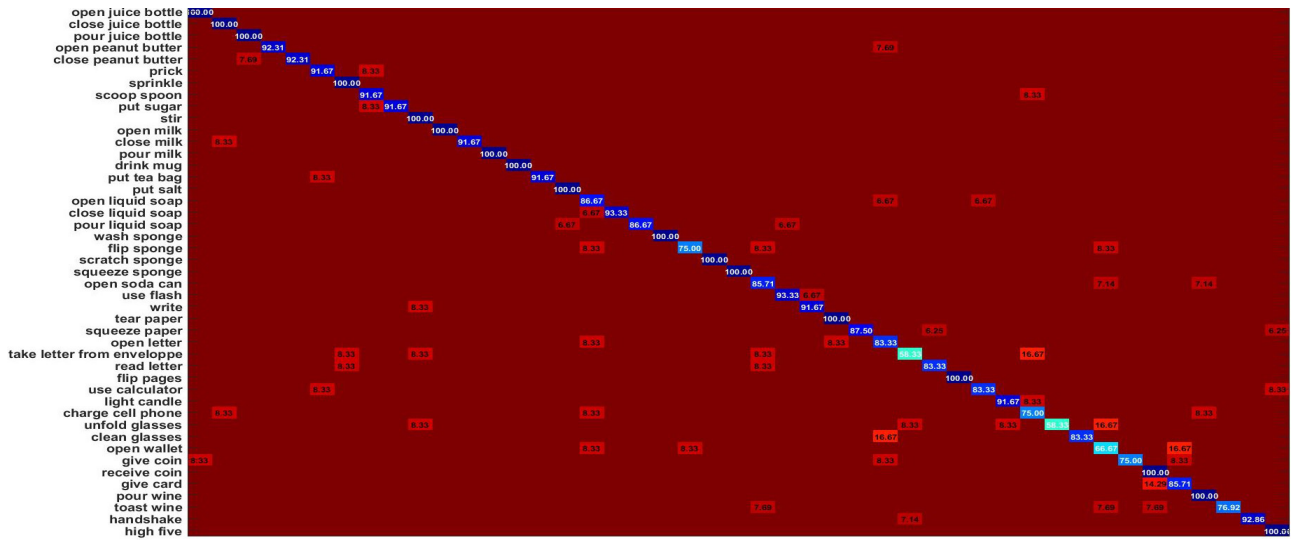}
\caption{Confusion matrix of the proposed network on the FPHA dataset.}
\label{fpha}
\end{figure*}

\subsection{Results on the DHG Dataset}
Results of the proposed method comparing with the existing hand-crafted feature based and deep learning based method \cite{oreifej2013hon4d,devanne20153,ohn2013joint,de2016skeleton, de2017shrec, devineau2018deep,nguyen2019a, avola2019exploiting, shin2020skeleton-based} are shown in Table~\ref{tab:resultsheet}. It can be seen that approaches based on deep learning generally outperform handcrafted feature based methods. Compared with methods based on CNNs~\cite{de2017shrec,devineau2018deep}, graph convolutional networks based method~\cite{li2019spatial} can capture spatial relationship of hand joints better and perform better. The proposed method outperforms these approaches, and achieves the best performance.

Fig.~\ref{rnn-cnn14} and Fig.~\ref{rnn-cnn28} show the confusion matrices on the DHG-14 and DHG-28 datasets, respectively, obtained with the proposed method. It can be seen that most hand gestures can be recognized effectively except hand gesture ``Grab (G)" and ``Pinch (P)". This is mostly because some Grab gestures only using the thumb and forefinger is of similar movement with ``Pinch". Therefore, networks to explore subtle differences are still highly desired and need to be investigated in the future. However, compared with results as illustrated in~\ref{tab:confusion14} of the previous work~\cite{shuai2020iscas}, our proposed method improves performance on all hand gestures.

\subsection{Results on the FPHA Dataset}
The comparison between the existing methods and the proposed method on the FPHA dataset is shown in Table~\ref{FPHAresult}. It also shows that our method achieves better performance than existing methods~\cite{feichtenhofer2016convolutional,hu2015jointly,garciahernando2017transition,zhang2016efficient,huang2018building}. The confusion matrix of FPHA dataset is illustrated in Fig.~\ref{fpha}. Most of the hand gestures are able to be accurately recognized. However, it is still difficult to recognize some gestures effectively such as ``open wallet", ``unfold glasses" and ``take letter from envelope" as shown in the confusion matrix. This is mostly because these hand gestures involve hand-object interaction, which cannot be well captured by the skeleton/pose alone.


\section{\textsc{Conclusion}}
\label{sec:CONCLUSION}
In this paper, a two-stream neural network is presented for recognizing the pose-based hand gesture. One is an SAGCN network having a self-attention mechanism to adaptively explore the collaboration among all joints in the spatial and short-term temporal domains. The other is a RBi-IndRNN to explore the long-term temporal dependency, compensating the weakness of SAGCN in processing the temporal features. The bidirectional processing and residual connections used in the RBi-IndRNN have proven to be effective in learning temporal patterns. State-of-the-art results are achieved by our two-stream neural network on two representative hand gesture datasets. Thorough analysis with ablation studies have also been conducted, validating the effectiveness of the proposed method.


\IEEEpeerreviewmaketitle


\bibliographystyle{IEEEtran}
\bibliography{IEEEtrans}

\begin{thebibliography}{10}
\providecommand{\url}[1]{#1}
\csname url@samestyle\endcsname
\providecommand{\newblock}{\relax}
\providecommand{\bibinfo}[2]{#2}
\providecommand{\BIBentrySTDinterwordspacing}{\spaceskip=0pt\relax}
\providecommand{\BIBentryALTinterwordstretchfactor}{4}
\providecommand{\BIBentryALTinterwordspacing}{\spaceskip=\fontdimen2\font plus
\BIBentryALTinterwordstretchfactor\fontdimen3\font minus
  \fontdimen4\font\relax}
\providecommand{\BIBforeignlanguage}[2]{{%
\expandafter\ifx\csname l@#1\endcsname\relax
\typeout{** WARNING: IEEEtran.bst: No hyphenation pattern has been}%
\typeout{** loaded for the language `#1'. Using the pattern for}%
\typeout{** the default language instead.}%
\else
\language=\csname l@#1\endcsname
\fi
#2}}
\providecommand{\BIBdecl}{\relax}
\BIBdecl

\bibitem{Cheng2019ARS}
L.~Cheng, Y.~Liu, Z.~Hou, M.~Tan, D.~Du, and M.~Fei, ``A rapid spiking neural
  network approach with an application on hand gesture recognition,''
  \emph{IEEE Transactions on Cognitive and Developmental Systems}, pp. 1--10,
  2019.

\bibitem{Xue2019MultimodalHH}
Y.~Xue, Z.~Ju, K.~Xiang, J.~Chen, and H.~Liu, ``Multimodal human hand motion
  sensing and analysis: A review,'' \emph{IEEE Transactions on Cognitive and
  Developmental Systems}, vol.~11, pp. 162--175, 2019.

\bibitem{Yang2020PerformanceCO}
Y.~Yang, F.~Duan, J.~Ren, J.~Xue, Y.~Lv, C.~Zhu, and H.~Yokoi, ``Performance
  comparison of gestures recognition system based on different classifiers,''
  \emph{IEEE Transactions on Cognitive and Developmental Systems}, pp. 1--10,
  2020.

\bibitem{7208833}
H.~{Cheng}, L.~{Yang}, and Z.~{Liu}, ``Survey on 3d hand gesture recognition,''
  \emph{IEEE Transactions on Circuits and Systems for Video Technology},
  vol.~26, no.~9, pp. 1659--1673, 2016.

\bibitem{Li2019DeepMS}
C.~Li, B.~Zhang, C.~Chen, Q.~Ye, J.~Han, G.~Guo, and R.~Ji, ``Deep manifold
  structure transfer for action recognition,'' \emph{IEEE Transactions on Image
  Processing}, vol.~28, pp. 4646--4658, 2019.

\bibitem{Cao2019SkeletonBasedAR}
C.~Cao, C.~Lan, Y.~Zhang, W.~Zeng, H.~Lu, and Y.~Zhang, ``Skeleton-based action
  recognition with gated convolutional neural networks,'' \emph{IEEE
  Transactions on Circuits and Systems for Video Technology}, vol.~29, pp.
  3247--3257, 2019.

\bibitem{Song2017AnES}
S.~Song, C.~Lan, J.~Xing, W.~Zeng, and J.~Liu, ``An end-to-end spatio-temporal
  attention model for human action recognition from skeleton data,'' in
  \emph{AAAI}, 2017.

\bibitem{de2016skeleton}
Q.~De~Smedt, H.~Wannous, and J.-P. Vandeborre, ``Skeleton-based dynamic hand
  gesture recognition,'' in \emph{IEEE Conference on Computer Vision and
  Pattern Recognition Workshops}, 2016, pp. 1--9.

\bibitem{de2017shrec}
Q.~De~Smedt, H.~Wannous, J.-P. Vandeborre, J.~Guerry, B.~Le~Saux, and
  D.~Filliat, ``Shrec'17 track: 3d hand gesture recognition using a depth and
  skeletal dataset,'' in \emph{Eurographics Workshop on 3D Object Retrieval},
  2017.

\bibitem{ohn2013joint}
E.~Ohn-Bar and M.~Trivedi, ``Joint angles similarities and hog2 for action
  recognition,'' in \emph{IEEE conference on computer vision and pattern
  recognition workshops}, 2013, pp. 465--470.

\bibitem{chen2017motion}
X.~Chen, H.~Guo, G.~Wang, and L.~Zhang, ``Motion feature augmented recurrent
  neural network for skeleton-based dynamic hand gesture recognition,'' in
  \emph{IEEE International Conference on Image Processing}, 2017, pp.
  2881--2885.

\bibitem{devineau2018deep}
G.~Devineau, F.~Moutarde, W.~Xi, and J.~Yang, ``Deep learning for hand gesture
  recognition on skeletal data,'' in \emph{IEEE International Conference on
  Automatic Face \& Gesture Recognition}, 2018, pp. 106--113.

\bibitem{devanne20153}
M.~Devanne, H.~Wannous, S.~Berretti, P.~Pala, M.~Daoudi, and A.~Del~Bimbo,
  ``3-d human action recognition by shape analysis of motion trajectories on
  riemannian manifold,'' \emph{IEEE transactions on cybernetics}, vol.~45,
  no.~7, pp. 1340--1352, 2015.

\bibitem{ren2011robust}
Z.~Ren, J.~Yuan, and Z.~Zhang, ``Robust hand gesture recognition based on
  finger-earth mover's distance with a commodity depth camera,'' in \emph{ACM
  international conference on Multimedia}, 2011, pp. 1093--1096.

\bibitem{Wang2012HandPR}
H.~Wang, Q.~Wang, and X.~Chen, ``Hand posture recognition from disparity cost
  map,'' in \emph{Asian Conference on Computer Vision}, 2012.

\bibitem{marin2014hand}
G.~Marin, F.~Dominio, and P.~Zanuttigh, ``Hand gesture recognition with leap
  motion and kinect devices,'' in \emph{IEEE International Conference on Image
  Processing}, 2014, pp. 1565--1569.

\bibitem{zhang2017learning}
L.~Zhang, G.~Zhu, P.~Shen, J.~Song, S.~A. Shah, and M.~Bennamoun, ``Learning
  spatiotemporal features using 3dcnn and convolutional lstm for gesture
  recognition,'' in \emph{IEEE Conference on Computer Vision and Pattern
  Recognition}, 2017, pp. 3120--3128.

\bibitem{shin2020skeleton-based}
S.~Shin and W.~Kim, ``Skeleton-based dynamic hand gesture recognition using a
  part-based gru-rnn for gesture-based interface,'' \emph{IEEE Access}, vol.~8,
  pp. 50\,236--50\,243, 2020.

\bibitem{narayana2018gesture}
P.~Narayana, J.~R. Beveridge, and B.~A. Draper, ``Gesture recognition: Focus on
  the hands,'' in \emph{IEEE Conference on Computer Vision and Pattern
  Recognition}, 2018, pp. 5235--5244.

\bibitem{molchanov2016online}
P.~Molchanov, X.~Yang, S.~Gupta, K.~Kim, S.~Tyree, and J.~Kautz, ``Online
  detection and classification of dynamic hand gestures with recurrent 3d
  convolutional neural network,'' in \emph{IEEE Conference on Computer Vision
  and Pattern Recognition}, 2016, pp. 4207--4215.

\bibitem{nguyen2019a}
X.~S. Nguyen, L.~Brun, O.~Lezoray, and S.~Bougleux, ``A neural network based on
  spd manifold learning for skeleton-based hand gesture recognition,'' in
  \emph{IEEE Conference on Computer Vision and Pattern Recognition}, 2019, pp.
  12\,036--12\,045.

\bibitem{li2019spatial}
Y.~Li, Z.~He, X.~Ye, Z.~He, and K.~Han, ``Spatial temporal graph convolutional
  networks for skeleton-based dynamic hand gesture recognition,'' \emph{Eurasip
  Journal on Image and Video Processing}, vol. 2019, no.~1, pp. 1--7, 2019.

\bibitem{Hu2020ProgressiveRL}
G.~Hu, B.~Cui, Y.~He, and S.~Yu, ``Progressive relation learning for group
  activity recognition,'' \emph{2020 IEEE/CVF Conference on Computer Vision and
  Pattern Recognition (CVPR)}, pp. 977--986, 2020.

\bibitem{li2018independently}
S.~Li, W.~Li, C.~Cook, C.~Zhu, and Y.~Gao, ``Independently recurrent neural
  network (indrnn): Building a longer and deeper rnn,'' in \emph{IEEE
  Conference on Computer Vision and Pattern Recognition}, 2018, pp. 5457--5466.

\bibitem{Li2019DeepIR}
S.~Li, W.~Li, C.~Cook, and Y.~Gao, ``Deep independently recurrent neural
  network (indrnn),'' \emph{ArXiv}, vol. abs/1910.06251, 2019.

\bibitem{FirstPersonAction_CVPR2018}
G.~Garcia-Hernando, S.~Yuan, S.~Baek, and T.-K. Kim, ``First-person hand action
  benchmark with rgb-d videos and 3d hand pose annotations,'' in \emph{IEEE
  Conference on Computer Vision and Pattern Recognition}, 2018.

\bibitem{shuai2020iscas}
S.~Li, L.~Zheng, C.~Zhu, and Y.~Gao, ``Bidirectional independently recurrent
  neural network for skeleton-based hand gesture recognition,'' in
  \emph{Proceedings of IEEE International Symposium on Circuits and Systems},
  2020.

\bibitem{Kuznetsova2013RealTimeSL}
A.~Kuznetsova, L.~Leal-Taix{\'e}, and B.~Rosenhahn, ``Real-time sign language
  recognition using a consumer depth camera,'' in \emph{2013 IEEE International
  Conference on Computer Vision Workshops}, 2013, pp. 83--90.

\bibitem{liu2017continuous}
Z.~Liu, X.~Chai, Z.~Liu, and X.~Chen, ``Continuous gesture recognition with
  hand-oriented spatiotemporal feature,'' in \emph{IEEE Conference on Computer
  Vision and Pattern Recognition}, 2017, pp. 3056--3064.

\bibitem{Guo2019AttentionBS}
S.~Guo, Y.~Lin, N.~Feng, C.~Song, and H.~Wan, ``Attention based
  spatial-temporal graph convolutional networks for traffic flow forecasting,''
  in \emph{AAAI Conference on Artificial Intelligence}, 2019.

\bibitem{Shi2019TwoStreamAG}
L.~Shi, Y.~Zhang, J.~Cheng, and H.~Lu, ``Two-stream adaptive graph
  convolutional networks for skeleton-based action recognition,'' in \emph{IEEE
  Conference on Computer Vision and Pattern Recognition}, 2019, pp.
  12\,018--12\,027.

\bibitem{yan2018spatial}
S.~Yan, Y.~Xiong, D.~Lin, and X.~Tang, ``Spatial temporal graph convolutional
  networks for skeleton-based action recognition,'' in \emph{AAAI Conference on
  Artificial Intelligence}, 2018, pp. 7444--7452.

\bibitem{liu2018multi}
X.~Liu, Y.~Li, and Q.~Wang, ``Multi-view hierarchical bidirectional recurrent
  neural network for depth video sequence based action recognition,''
  \emph{International Journal of Pattern Recognition and Artificial
  Intelligence}, p. 1850033, 2018.

\bibitem{arisoy2015bidirectional}
E.~Arisoy, A.~Sethy, B.~Ramabhadran, and S.~Chen, ``Bidirectional recurrent
  neural network language models for automatic speech recognition,'' in
  \emph{IEEE International Conference on Acoustics, Speech and Signal
  Processing}, 2015, pp. 5421--5425.

\bibitem{kingma2014adam}
D.~P. Kingma and J.~Ba, ``Adam: A method for stochastic optimization,''
  \emph{arXiv preprint arXiv:1412.6980}, 2014.

\bibitem{srivastava2014dropout}
N.~Srivastava, G.~Hinton, A.~Krizhevsky, I.~Sutskever, and R.~Salakhutdinov,
  ``Dropout: a simple way to prevent neural networks from overfitting,''
  \emph{The Journal of Machine Learning Research}, vol.~15, no.~1, pp.
  1929--1958, 2014.

\bibitem{oreifej2013hon4d}
O.~Oreifej and Z.~Liu, ``Hon4d: Histogram of oriented 4d normals for activity
  recognition from depth sequences,'' in \emph{IEEE conference on computer
  vision and pattern recognition}, 2013, pp. 716--723.

\bibitem{avola2019exploiting}
D.~Avola, M.~Bernardi, L.~Cinque, G.~L. Foresti, and C.~Massaroni, ``Exploiting
  recurrent neural networks and leap motion controller for the recognition of
  sign language and semaphoric hand gestures,'' \emph{IEEE Transactions on
  Multimedia}, vol.~21, no.~1, pp. 234--245, 2019.

\bibitem{feichtenhofer2016convolutional}
C.~Feichtenhofer, A.~Pinz, and A.~Zisserman, ``Convolutional two-stream network
  fusion for video action recognition,'' in \emph{IEEE Conference on Computer
  Vision and Pattern Recognition}, 2016, pp. 1933--1941.

\bibitem{hu2015jointly}
J.~Hu, W.-S. Zheng, J.-H. Lai, and J.~Zhang, ``Jointly learning heterogeneous
  features for rgb-d activity recognition,'' \emph{IEEE Transactions on Pattern
  Analysis and Machine Intelligence}, vol.~39, pp. 2186--2200, 2017.

\bibitem{rahmani20163d}
H.~Rahmani and A.~Mian, ``3d action recognition from novel viewpoints,'' in
  \emph{IEEE Conference on Computer Vision and Pattern Recognition}, 2016, pp.
  1506--1515.

\bibitem{zhu2016co-occurrence}
W.~Zhu, C.~Lan, J.~Xing, W.~Zeng, Y.~Li, L.~Shen, and X.~Xie, ``Co-occurrence
  feature learning for skeleton based action recognition using regularized deep
  lstm networks,'' in \emph{IEEE Conference on Computer Vision and Pattern
  Recognition}, 2016, pp. 3697--3703.

\bibitem{zanfir2013the}
M.~Zanfir, M.~Leordeanu, and C.~Sminchisescu, ``The moving pose: An efficient
  3d kinematics descriptor for low-latency action recognition and detection,''
  in \emph{IEEE International Conference on Computer Vision}, 2013, pp.
  2752--2759.

\bibitem{vemulapalli2014human}
R.~Vemulapalli, F.~Arrate, and R.~Chellappa, ``Human action recognition by
  representing 3d skeletons as points in a lie group,'' in \emph{IEEE
  Conference on Computer Vision and Pattern Recognition}, 2014, pp. 588--595.

\bibitem{du2015hierarchical}
Y.~Du, W.~Wang, and L.~Wang, ``Hierarchical recurrent neural network for
  skeleton based action recognition,'' in \emph{IEEE Conference on Computer
  Vision and Pattern Recognition}, 2015, pp. 1110--1118.

\bibitem{zhang2016efficient}
X.~Zhang, Y.~Wang, M.~Gou, M.~Sznaier, and O.~Camps, ``Efficient temporal
  sequence comparison and classification using gram matrix embeddings on a
  riemannian manifold,'' in \emph{IEEE Conference on Computer Vision and
  Pattern Recognition}, 2016, pp. 4498--4507.

\bibitem{garciahernando2017transition}
G.~Garciahernando and T.~Kim, ``Transition forests: Learning discriminative
  temporal transitions for action recognition and detection,'' in \emph{IEEE
  Conference on Computer Vision and Pattern Recognition}, 2017, pp. 407--415.

\bibitem{huang2016a}
Z.~Huang and L.~Van~Gool, ``A riemannian network for spd matrix learning,'' in
  \emph{AAAI Conference on Artificial Intelligence}, 2016, pp. 2036--2042.

\bibitem{huang2018building}
Z.~Huang, J.~Wu, and L.~Van~Gool, ``Building deep networks on grassmann
  manifolds,'' in \emph{AAAI Conference on Artificial Intelligence}, 2018, pp.
  3279--3286.

\end{thebibliography}

\ifCLASSOPTIONcaptionsoff
  \newpage
\fi

\begin{IEEEbiography}[{\includegraphics[width=1in,height=1.25in,clip,keepaspectratio]{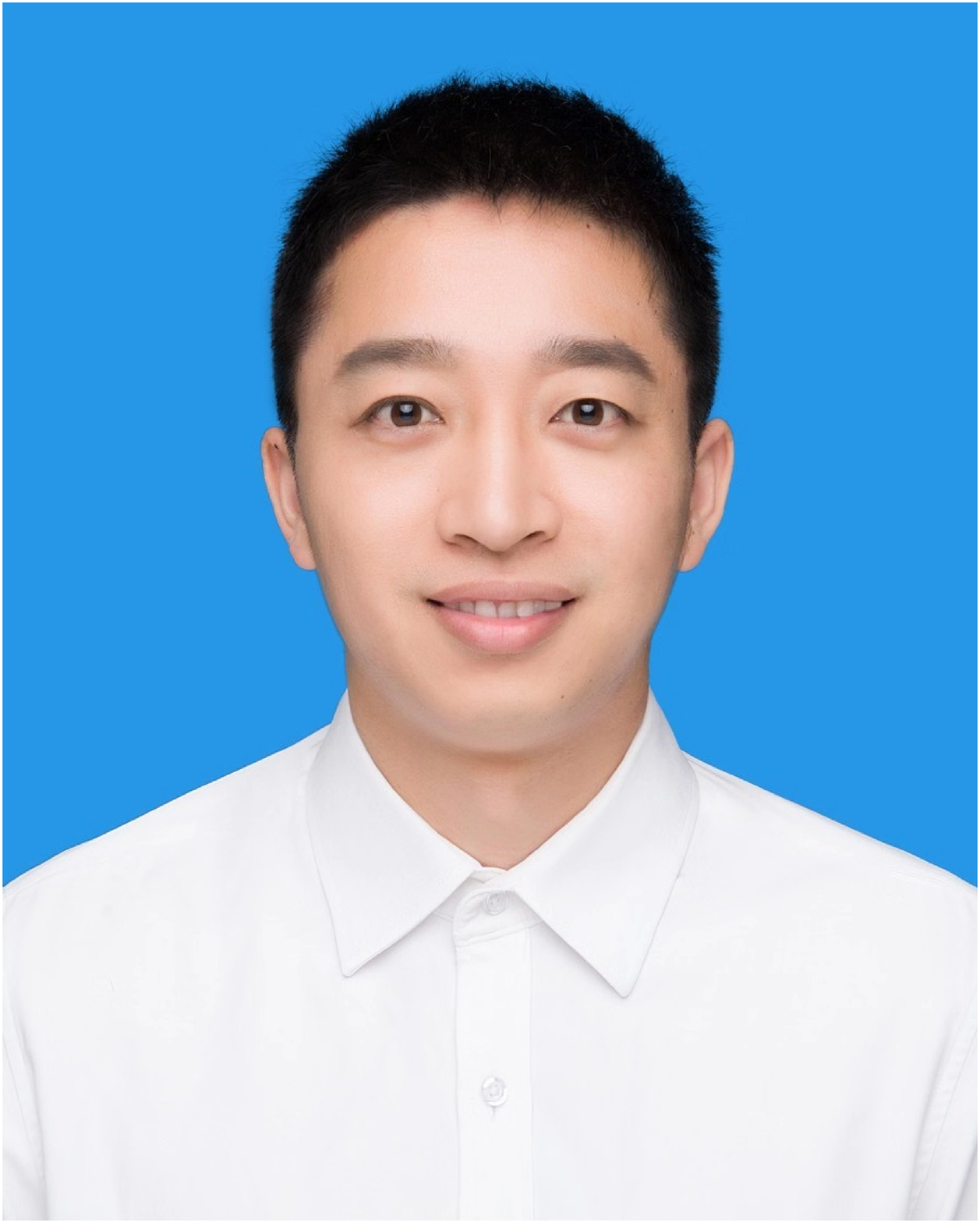}}]{Chuankun Li}
received the BE degree in electronic information engineering from North University of China, Taiyuan, China, in 2012 and received the MS degree in communication and information system from North University of China, Taiyuan, China, in 2015. He received the Ph.D degree with School of electronic information engineering , Tianjin University, China in 2020. His current research interests include computer vision and machine learning.
\end{IEEEbiography}

\begin{IEEEbiography}[{\includegraphics[width=1in,height=1.25in,clip,keepaspectratio]{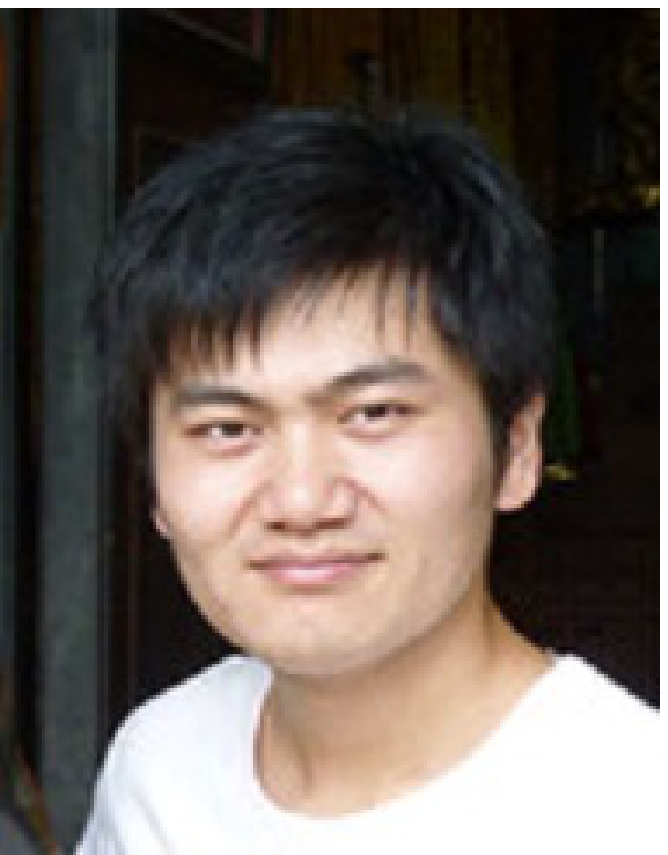}}]{Shuai Li}
is currently with the School of Control Science and Engineering, ShanDong University (SDU), China, as a Professor and QiLu Young Scholar. He was with the School of Information and Communication Engineering, University of Electronic Science and Technology of China, China, as an Associate Professor from 2018-2020. He received his Ph.D. degree from the University of Wollongong, Australia, in 2018. His research interests include image/video coding, 3D video processing and computer vision. He was a co-recipient of two best paper awards at the IEEE BMSB 2014 and IIH-MSP 2013, respectively.
\end{IEEEbiography}

\begin{IEEEbiography}[{\includegraphics[width=1in,height=1.25in,clip,keepaspectratio]{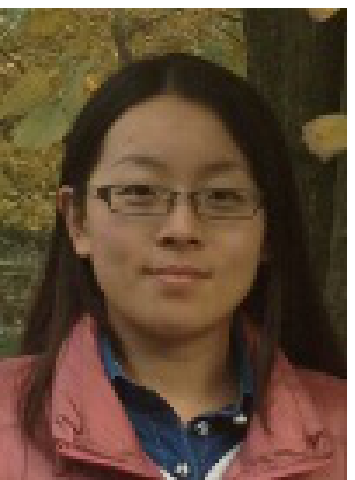}}]{Yanbo Gao}
is currently with the School of Software, Shandong University (SDU), Jinan, China, as an Associate Professor. She was with the School of Information and Communication Engineering, University of Electronic Science and Technology of China (UESTC), Chengdu, China, as a Post-doctor from 2018-2020. She received her Ph.D. degree from UESTC in 2018. Her research interests include video coding, 3D video processing and light field image coding. She was a co-recipient of the best student paper awards at the IEEE BMSB 2018.
\end{IEEEbiography}

\begin{IEEEbiography}[{\includegraphics[width=1in,height=1.25in,clip,keepaspectratio]{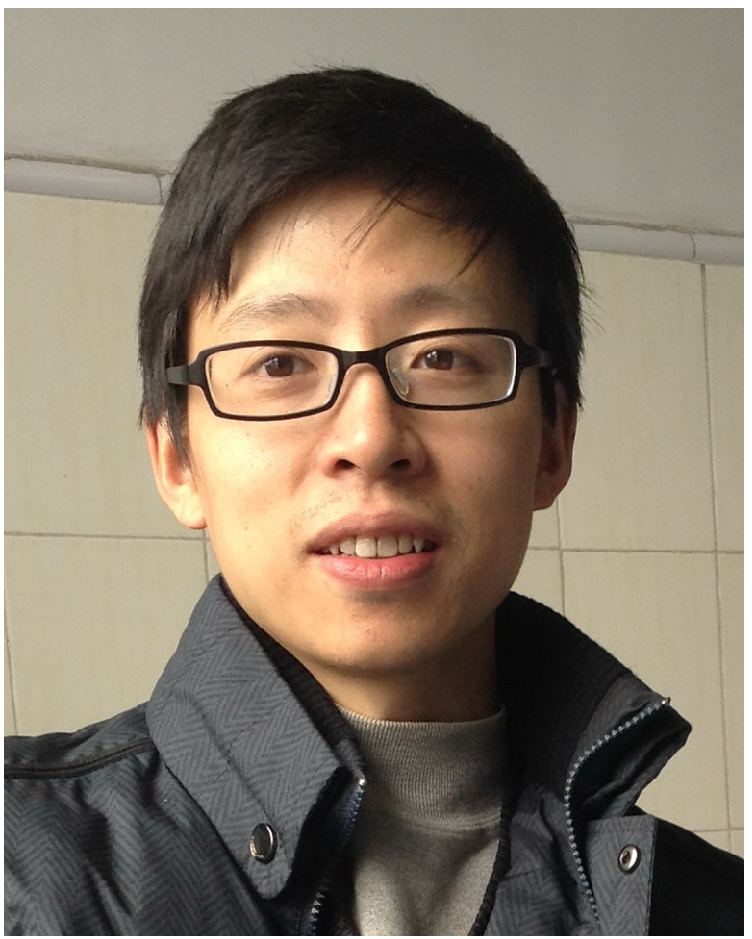}}]{Xiang Zhang}
received the B.S. and M.S. degrees from University of Electronic Science and Technology of China, Chengdu, China, and the Ph.D. degree from Shanghai Jiaotong University, Shanghai, China, in 2003, 2006, and 2009, respectively. He is an Associate Professor with the School of Information and Communication Engineering, University of Electronic Science and Technology of China. His research interests include video analysis and machine learning.
\end{IEEEbiography}

\begin{IEEEbiography}[{\includegraphics[width=1in,height=1.25in,clip,keepaspectratio]{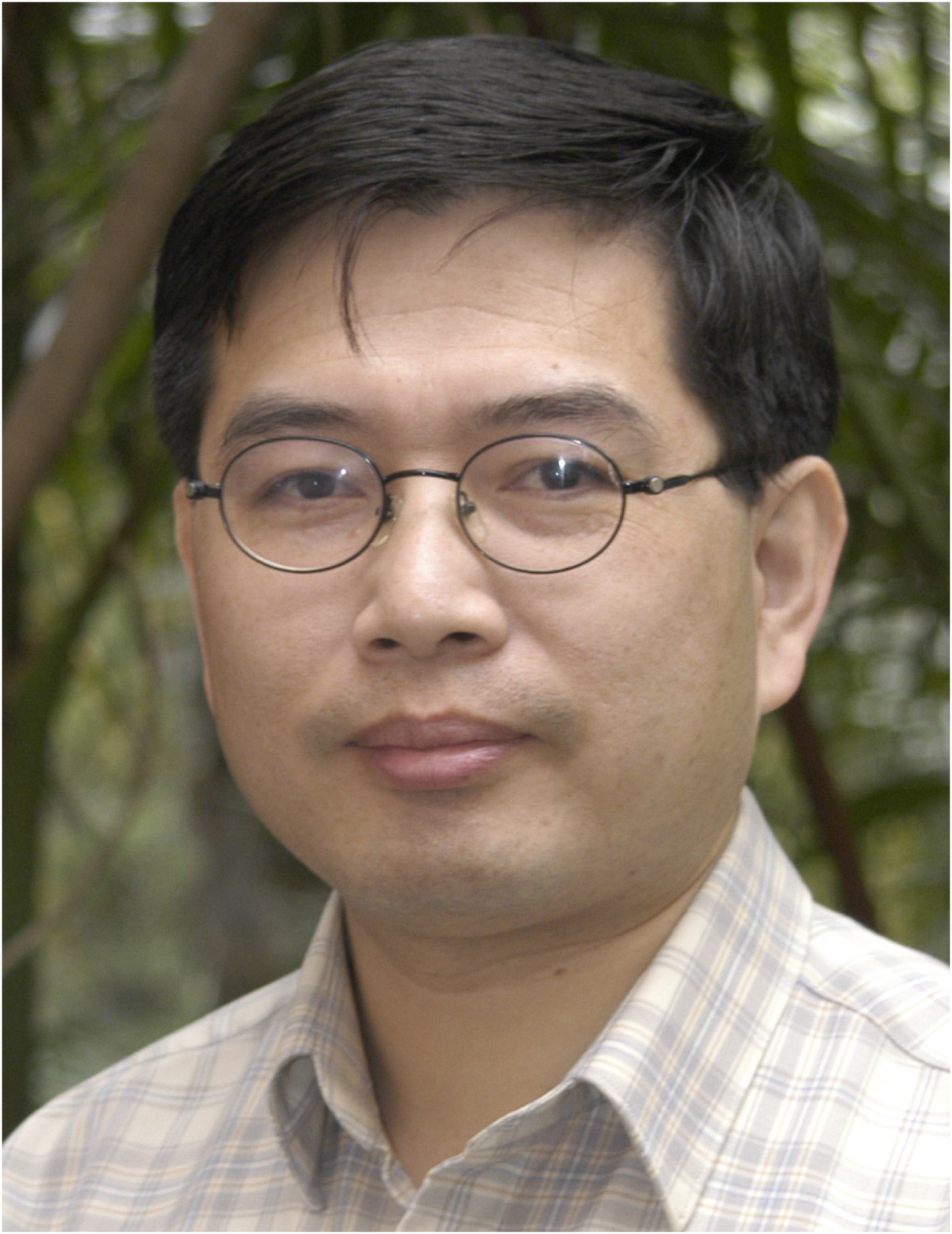}}]
{Wanqing Li} (M'97-SM'05) received his Ph.D. in electronic engineering from The University of Western Australia. He was a Principal Researcher (98-03) at Motorola Research Lab and a visiting researcher (08, 10 and 13) at Microsoft Research US. He is currently an Associate Professor and Co-Director of Advanced Multimedia Research Lab (AMRL) of UOW, Australia. His research areas are machine learning, 3D computer vision, 3D multimedia signal processing and medical image analysis. Dr. Li currently serves as an Associate Editor for \textsc{IEEE Transactions on Circuits and Systems for Video Technology} and \textsc{IEEE Transactions on Multimedia}. He was an Associator for \textsc{Journal of Visual Communication and Image Representation}.
\end{IEEEbiography}
\vfill






\end{document}